\documentclass[letterpaper]{article} %
\usepackage{aaai2026}  %
\usepackage{times}  %
\usepackage{helvet}  %
\usepackage{courier}  %
\usepackage[hyphens]{url}  %
\usepackage{graphicx} %
\usepackage{natbib}  %
\usepackage{caption} %
\usepackage{algorithm}
\usepackage{algorithmic}

\usepackage{newfloat}
\usepackage{listings}
\DeclareCaptionStyle{ruled}{labelfont=normalfont,labelsep=colon,strut=off} %
\floatstyle{ruled}
\newfloat{listing}{tb}{lst}{}
\floatname{listing}{Listing}
\usepackage{soul}
\usepackage{todonotes}
\usepackage{amsmath}
\usepackage{enumitem}
\usepackage{placeins}
\usepackage{csquotes}
\usepackage{tikz}
\usepackage{fancyvrb}
\usepackage{dsfont}
\usepackage{booktabs}
\usepackage{multirow}

\usepackage{soul}
\usepackage{xspace}
\usepackage[table]{xcolor}
\usepackage{array}

\setlist[description]{leftmargin=\parindent,labelindent=0pt,itemsep=0pt,topsep=0pt,partopsep=0ex,parsep=0ex}

\definecolor{darkgreen}{rgb}{0.0, 0.5, 0.0}
\definecolor{amaranth}{rgb}{0.9, 0.17, 0.31}

\definecolor{PineGreen}{rgb}{0.0, 0.47, 0.44}

\usepackage{amssymb}
\usepackage{utfsym}

\usepackage{csquotes}

\usepackage{pifont}%

\usepackage{csquotes}

\title{A Solver-in-the-Loop Framework for Improving LLMs\\ on Answer Set Programming for Logic Puzzle Solving}
\author {
    Timo Pierre Schrader\textsuperscript{\rm 1,\rm 2}, 
    Lukas Lange\textsuperscript{\rm 1}, 
    Tobias Kaminski\textsuperscript{\rm 1}, \\
    Simon Razniewski\textsuperscript{\rm 3}, 
    Annemarie Friedrich\textsuperscript{\rm 2}
}
\affiliations {
    \textsuperscript{\rm 1}Bosch Center for AI, Renningen, Germany \\
    \textsuperscript{\rm 2}University of Augsburg, Germany\\
    \textsuperscript{\rm 3}ScaDS.AI \& TU Dresden, Germany\\
    timo.schrader@de.bosch.com
}

\usepackage{bibentry}

\begin{document}

\maketitle

\begin{abstract}
The rise of large language models (LLMs) has sparked interest in coding assistants.
While general-purpose programming languages are well supported, generating code for domain-specific languages remains a challenging problem for LLMs.
In this paper, we focus on the LLM-based generation of code for Answer Set Programming (ASP), a particularly effective approach for finding solutions to combinatorial search problems.
The effectiveness of LLMs in ASP code generation is currently hindered by the limited number of examples seen during their initial pre-training phase.

In this paper, we introduce a novel ASP-solver-in-the-loop approach for solver-guided instruction-tuning of LLMs to addressing the highly complex semantic parsing task inherent in ASP code generation.
Our method only requires problem specifications in natural language and their solutions.
Specifically, we sample ASP statements for program continuations from LLMs for unriddling logic puzzles.
Leveraging the special property of declarative ASP programming that partial encodings increasingly narrow down the solution space, we categorize them into chosen and rejected instances based on solver feedback.
We then apply supervised fine-tuning to train LLMs on the curated data and further improve robustness using a solver-guided search that includes best-of-N sampling.
Our experiments demonstrate consistent improvements in two distinct prompting settings on two datasets.
\end{abstract}

\begin{links}
    \link{Code}{https://bos.ch/1qzx1jj}
\end{links}

\section{Introduction}
\label{sec:one}
One of the primary productive applications of large language models (LLMs) is currently their use as coding assistants \citep{jiang2024surveylargelanguagemodels,gu2023llm,ugare2024improving}, supporting software developers by taking over tedious and repetitive programming workflows.
However, current systems typically target popular and general-purpose programming languages such as JavaScript, C++, and Python.

Many real-life problems, however, require the use of domain-specific programming languages tailored towards specific problem types.
Various studies on LLM-based code generation have been conducted on programming languages such as PROLOG or PDDL (\textit{planning domain definition language}, \citet{McDermott1998PDDLthePD}). Existing approaches rely on prompt engineering or create complex planning and reasoning systems \citep{yang2024arithmetic,stein2023autoplanbench,vyas2024llm}.
Yet, for many programming languages, LLM-based generations remain understudied.

In this work, we focus on \textit{Answer Set Programming} \citep[ASP,][]{gelfond1988stable,marek1999stable}, which can be used for finding solutions to combinatorial search problems.
ASP is a well-known and powerful approach for declarative problem solving. It offers an expressive input language, provides well-optimized solvers, and has been applied to a wide range of industrial problems such as generating software test cases, robotics, and configuration problems \citep{falkner2018industrial}.

\begin{figure*}[!t]
    \centering
    \includegraphics[width=1.0\linewidth]{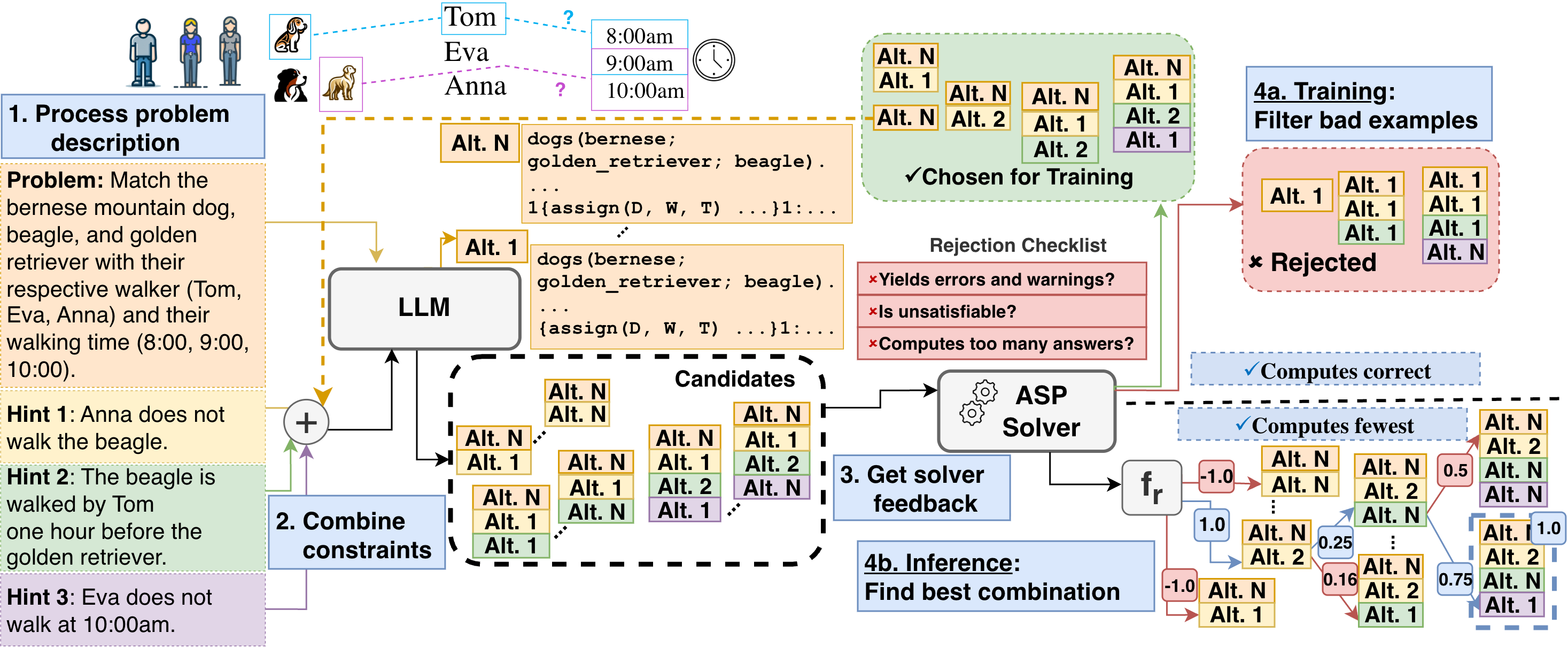}
    \caption{
    We use the feedback from an ASP solver to assess LLM-generated ASP statements. \textbf{(1)} We take the problem description and let the LLM output $N$ alternatives for it (\textbf{generate} part of an ASP encoding). We then judge the correctness of these encodings using an ASP solver which checks for errors as well as for the expected answer size. \textbf{(2)} We incrementally combine the constraints of the problem into larger inputs to the LLM and obtain longer partial encodings (\textbf{check} part of an ASP encoding). \textbf{(3)} The ASP solver judges the correctness of each partial encoding containing the base encoding and at least one constraint. \textbf{(4a)} For training, we select instances from a chosen set that corresponds to instances with positive solver feedback. \textbf{(4b)} During inference, where the ground truth is not known, we score every partial encoding using a reward function that captures errors as well as the number of produced answer sets with preference for a lower number of returned answers. Additionally, we implement fallback mechanisms that regenerate encodings or backtrack in the graph if all alternatives were judged with a negative score.}
    \label{fig:teaser}
\end{figure*}

ASP is highly suited for solving many real-life problems, such as scheduling in factory plants, configuring electrical circuits (ECUs), or assignment problems as shown in Figure \ref{fig:teaser}.
These problems and their constraints can often be described by domain experts in natural language.
However, domain specialists often lack the necessary ASP programming skills. 
Out-of-the-box LLMs frequently struggle to solve these problems using chain-of-thought reasoning, particularly as problem size and complexity grow.
Neuro-symbolic solutions that combine LLMs and symbolic solvers are a promising alternative \cite{olausson-etal-2023-linc,schrader-etal-2024-quite}.
These systems translate fine-grained natural language statements into executable and semantically correct code for which a solver computes the solution.

First explorations addressing ASP code generation include the LLASP dataset \citep{coppolillo2024llasp} that provides ASP statements only in isolation without a larger contextual problem.
\citet{ishay2023leveraging} engineer a highly specific prompt pipeline for a particular dataset for commercial GPT-based models only.
For processing sensitive data on-premise, the use of open-weight LLMs is an alternative.
\textbf{Yet, as we show in this paper, state-of-the-art open-weight LLMs still mostly fall short on producing correct ASP encodings, even when being provided with strong prompt guidance.}
Our goal is to improve trainable state-of-the-art LLMs on the task of ASP code generation.

In this paper, we present a novel and efficient method that uses an ASP solver in the loop both for creating high-quality training data and for guiding the generation process with a solver-based reward.
Our ultimate goal is to develop LLMs that excel at ASP coding for assignment problems.
We focus on solving grid-based puzzles as a proxy task since they reflect assignment problems that are one of the main use cases of ASP.
The LLM-generated ASP encodings are automatically evaluated by an ASP solver and classified as either \textit{chosen} or \textit{rejected}.
Our solver-in-the-loop setup leverages the specific properties of ASP as a declarative approach, i.e., that the order in which rules and constraints are added does not alter the semantics of the final encoding. Hence, \textbf{it allows for fine-grained evaluation of partial encodings regardless of the order in which they are added by checking if the ground truth solution can still be derived from the partial encoding.}
We also make use of the fact that ASP encodings are usually divided into clearly defined \textit{generate} and \textit{check} parts.
This enables us to (re-)combine partial ASP encodings and to verify intermediate generation steps by checking for both syntax errors as well as for a reduction of the answer space after each added constraint.
This specific ASP property allows for a substantially more fine-grained evaluation than evaluating code of other programming languages, e.g., in Python, semantic feedback from the interpreter can only be obtained for a full block without being able to check intermediate steps \citep{peng2024perfcodegenimprovingperformancellm}.

Using the chosen continuations, we apply supervised fine-tuning (SFT) via causal language modeling. %
Additionally, we leverage the same solver-in-the-loop setup to further improve the ASP code generation during test-time.
We introduce a novel reward function that leverages solver feedback to rank a set of N alternatives for the same input and triggers fallback mechanisms in case of errors.

We evaluate our trained models with and without reward-based inference for solving logic puzzles in two distinct settings:
First, in a \textbf{few-shot setting}, we only provide basic instructions as well as a few examples in the input prompt. This reflects a realistic setting in which the type of problem and dataset format is not clearly defined beforehand.
Second, a \textbf{prompt-engineered setting} \citep{ishay2023leveraging} assumes that everything about the input problem and how it could be generally modeled in ASP is already fully known.

Our contributions are as follows:
(1) We demonstrate that state-of-the-art open-weight LLMs struggle with ASP code generation, even under highly prompt-engineered conditions.
(2) We create silver-standard training data by sampling partial ASP encodings from an LLM for given combinatorial problems and filtering them based on feedback from an ASP solver that includes whether the encoding is erroneous or whether the ground truth answer can be derived from it.
(3) We show that applying SFT on the chosen instances %
improves the performance of open-weight LLMs of various sizes in generating ASP encodings for grid-based puzzles, especially for problems of higher complexity in both prompt settings.
(4) We demonstrate that the solver can steer the ASP generation process successfully during inference.

\section{Answer Set Programming}
\label{sec:three}
ASP, a form of declarative programming focusing on difficult, primarily NP-hard search problems, is based on the stable model semantics proposed by \citet{gelfond1988stable}. 
Our short introduction follows \citet{lifschitz2008answer}.
The purpose of ASP is to find answer sets consisting of sets of instantiated first-order atoms (e.g., \texttt{walk(beagle, eva, 8)}) that satisfy all given rules and constraints.
We exemplify ASP with the puzzle of matching dogs with their walkers and times for walking (cf. Figure~\ref{fig:teaser}).

\textbf{Rules} define if-then statements written in PROLOG-style syntax using $n$-ary predicates.
The right-hand side (\textit{rule body})
is the premise and the left-hand side the conclusion (\textit{rule head}).
\enquote{If the beagle is walked by Tom, then the golden retriever will be walked by Eva at 9 am.} is encoded as:

\begin{small}
\begin{Verbatim}[frame=single,framesep=1pt]
  walk(golden, eva, 9) :- 
               walk(beagle, tom, _).
\end{Verbatim}
\end{small}
\textbf{Constraints} eliminate undesired solutions by disallowing certain combinations of atoms to be true simultaneously.
They only have a rule body, i.e., no head.
A constraint requires at least 1 atom to evaluate to \texttt{false}. 
For example, \enquote{Anna walks her dog later than Eva.} is encoded as:

\begin{footnotesize}
\begin{Verbatim}[frame=single,framesep=1pt]
 :- walk(_, anna, T1), walk(_, eva, T2), 
    not T1 > T2.
\end{Verbatim}
\end{footnotesize}
Unlike constraints, \textbf{choice rules} generate answer set candidates instead of filtering them.
Assuming that the ASP program already contains atoms specifying a set of entities (here: dogs, persons, and times), the following rule states that \enquote{For each time T, some dog D is walked by some person P.}
The statement \texttt{dog(beagle;golden;bernese)} is an ASP shorthand for \texttt{dog(beagle). dog(golden). dog(bernese)}.

\begin{footnotesize}
\begin{Verbatim}[frame=single]
 person(tom;eva;anna).
 dog(beagle;golden;bernese).
 time(8;9;10).
    
 1 { walk(D, P, T) : dog(D), person(P) } 1 
    :- time(T).
\end{Verbatim}
\end{footnotesize}

\noindent\textbf{Solutions in ASP.}
An ASP solver calculates all answer sets, corresponding to minimal sets of derivable atoms that are valid interpretations of all rules.
More technically, a solution of an ASP program $\Pi$ is a special truth assignment to all atoms that occur in the grounding of the program, i.e., a stable model, and the effect of adding a constraint to a program can lead to its elimination, i.e., constraints reduce the set of solutions monotonically.

\section{Instruction Tuning for ASP} 

\label{sec:four}
We present a novel instruction-tuning method for ASP code generation in LLMs.
We prompt an LLM $\mathcal{M}_S$ to generate ASP code and classify the results into \enquote{chosen} and \enquote{rejected} samples using an ASP solver.
The chosen data can be used with causal language modeling to train an ASP-specific model $\mathcal{M}_{ASP}$ based on the reference model $\mathcal{M}_S$. %

\subsection{Task Definition}
\label{sec:task_definition}

The problems addressed in this paper are grid-based puzzle instances of the form $\mathcal{I} = \{\mathcal{D}, \mathcal{E}, \mathcal{H}, \mathcal{S}\}$.
$\mathcal{D}$ refers to the natural language problem description, $\mathcal{E}$ to a set of entities and their types ({%
e.g., dogs, persons, and times}), 
$\mathcal{H}$ to a set of natural language hints (or clues) that constrain the answer space 
(\enquote{Clue: The beagle is walked one hour before the poodle.}), 
and $\mathcal{S}$ refers to the correct solution assignment.
To solve this, the LLM $\mathcal{M}_{\text{ASP}}$ has to parse the input problem  $\{\mathcal{D}, \mathcal{E}, \mathcal{H}$\} specified as natural language text into a valid ASP encoding $\Pi$ that contains ASP encodings for entities, (choice) rules, and constraints.
A solver will then compute the solution $\mathcal{S}$ given $\Pi$.
In the following, we stick to the notations of \citet{lifschitz2008answer}, using $\Gamma$ to denote a partial ASP encoding that contains a proper subset of the ASP code in $\Pi$, i.e., $solution(\Pi = (\Gamma_1\cup\dots\cup\Gamma_n)) = \mathcal{S}$.

\subsection{Sampling Trajectories}
We define a \textit{trajectory} $\mathcal{T}$ to be an alternating sequence of natural language inputs and ASP statements.
A trajectory $\mathcal{T}$ starts with a prompt $P_{\mathcal{D},\mathcal{E}}$ comprising general information on ASP,\footnote{This includes an instruction of how different variants of logical OR works in ASP, a hint to introduce helper predicates to perform arithmetics (e.g., on dates), a high-level explanation of the steps required to solve a grid-based puzzle in ASP as well as a basic choice rule.} 
the textual problem description $\mathcal{D}$, and the set of entities $\mathcal{E}$. %
We feed $\mathcal{P}_{\mathcal{D},\mathcal{E}}$
into the LLM $\mathcal{M}_S$ to obtain ASP code $\Gamma_{\mathcal{C},\mathcal{E}}$, where $\mathcal{C}$ refers to the choice rule that initially generates all potential solutions representing the problem instance, which is then appended to the trajectory. 
Next, a natural language hint $h_i \in \mathcal{H}$ is appended to this trajectory and $\mathcal{M}_S$ is prompted again to obtain a partial encoding $\Gamma_{i}$.
This results in trajectories of form $\mathcal{T} = \{\mathcal{P_{\mathcal{D},\mathcal{E}}} , \Gamma_{\mathcal{C},\mathcal{E}}, h_1, \Gamma_{1}, h_2, \Gamma_{2}, \dots, h_n, \Gamma_{n}\}$ with $n$ being the number of hints of the problem.

\subsection{Classification of ASP Encodings}
After processing $k$ of the $n$ hints, we obtain a trajectory $\mathcal{T}_{k}$ that contains $k$+1 ASP encodings (one for entities and choice rule and $k$ for hints). 
We now combine all $k$+1 ASP encodings into the partial encoding $\Gamma_{\mathcal{C},\mathcal{E}}\cup\Gamma_1\cup\dots\Gamma_k$ 
and use the solver to compute its solution.
We sample 5 completions from $\mathcal{M}_S$ for each step $k$ simultaneously to get a diverse set of ASP encodings for the same input. Preliminary experiments showed that a temperature $t = 0.8$ for sampling provides a good balance between chosen and rejected responses.
We then use an ASP solver to evaluate if the partial encoding $\Gamma_k$ generates a solution that comprises the ground truth answer set $\mathcal{S}$.
We consider $\Gamma_k$ as a \textit{chosen} response to input $\mathcal{T}_k = \{\mathcal{P_{\mathcal{D},\mathcal{E}}} , \Gamma_{\mathcal{C},\mathcal{E}}, h_1, \Gamma_{1}, h_2, \Gamma_{2}, \dots ,h_k\}$ if the solution of the partial program $\Gamma_{\mathcal{C},\mathcal{E}}\cup\Gamma_1\cup\dots\Gamma_k$ comprises the ground truth solution.
If it produces only wrong answer sets, becomes unsatisfiable, or errors and warnings are returned by the solver, %
we consider $\Gamma_k$ as \textit{rejected}. 

\subsection{Training Data Generation}
We generate trajectories using depth-first search. 
At each step $k$, we consider at most two chosen responses.
For each of them, we continue at step $k$+1 recursively until there are no hints left.
To form pairs from both chosen and rejected responses at each step $k$, we take the Cartesian product between both sets.
To put emphasis on instances that are difficult for current LLMs, we keep only instances for which there are at least one chosen and one rejected alternative.
If we have a mix of chosen and rejected responses, we obtain either $1 \times 4$ or $2 \times 3$ preference pairs in each step.
If there are only chosen or only rejected responses, the trajectory generation is continued at the next step without creating preference pairs. 
If all responses are rejected, the input to the next step becomes $\{\mathcal{P_{\mathcal{D},\mathcal{E}}} , \Gamma_{\mathcal{C},\mathcal{E}}, h_1, \Gamma_{1},  \ldots,h_{k-1}, \Gamma_{k-1}, h_{k+1} \}$, i.e., the rejected response and the hint prompt causing it are removed. This is possible because the hints in grid-based puzzles do not refer to each other.

\subsection{Model Training}
The chosen instances can be used to train LLMs using SFT with causal language modeling.
Alternative training methods, such as DPO, could also be used to perform preference alignment using pairs of chosen and rejected instances.
However, our initial experiments showed that SFT works slightly better than DPO in this context.

\section{Test-Time Sampling for ASP}
\label{sec:five}

In this section, we explain our best-of-N sampling method that can be used during test-time on trained and untrained LLMs.
It is based on a novel solver-grounded reward function for LLMs that helps to generate correct ASP encodings more reliably.

\subsection{Reward Function}
\label{ssec:reward_function}

The reward function $f_r$ maps an encoding $\Gamma$ and the number $M \in \mathbb{N}$ of answer sets produced by $\Gamma$ to a floating point reward $r \in R$ that aims to judge the quality of $\Gamma$:

{\small
\begin{equation*}
    f_r(\Gamma, M) = \frac{1}{M} - \mathds{1}_E(\Gamma) - \mathds{1}_U(\Gamma) - \mathds{1}_{NE}(\Gamma)
\end{equation*}
}

Usually, with every hint in a logic puzzle, the number of possible answers is either reduced or stays the same. %
Therefore, $f_r$ rewards stricter generations.
$\mathds{1}$ are binary indicator variables checking $\Gamma$ for undesired properties. All three contribute negatively to the reward:

\hangindent=2em \hangafter=1 $\mathds{1}_E$ indicates whether there are any errors or warnings when trying to solve $\Gamma$.

\hangindent=2em \hangafter=1 $\mathds{1}_U$ refers to unsatifiability, i.e., there is no warning or error, but also no answer set.

\hangindent=2em \hangafter=1 $\mathds{1}_{NE}$ indicates that a manually specified maximum number of answer sets is exceeded.
This avoids out-of-memory issues in cases where broken encodings lead to combinatorial explosions. %

\subsection{Sampling Procedure}

Our test-time method is based on \textit{greedy} search, i.e., it aims at maximizing the current reward and does not perform trade-offs in favor of future rewards (cf. \citet{sutton1998reinforcement}).
Similar to creating ASP encodings for training, we first instruct $\mathcal{M}_{ASP}$ to generate $N$ alternatives for the partial encoding $\Gamma_{\mathcal{C,E}}$ that encodes entities and the choice rule.
We then select and keep the alternative receiving the highest reward according to $f_r$.
We use a special version of $f_r$ for evaluating the $\Gamma_{\mathcal{C,E}}$ encodings, by replacing the reciprocal factor $\frac{1}{M}$ with a strict check whether the output contains $(n!)^{m-1}$ answer sets for an $m \times n$ grid puzzle which is different from $\mathds{1}_{NE}$ in that it is dynamically determined based on the size of the input problem.
This corresponds to the number of all theoretically possible answer combinations when disregarding the constraints. %
Next, for every hint $h_j \in \mathcal{H}$ in sequential order, we generate $N$ alternatives and append them to the partial encoding that contains all previously selected encodings and judge all $N$ partial encodings again by $f_r$.
At each step, we select the partial encoding with the highest score and continue until we arrive at the full encoding $\Pi$.
We make use of two recovery mechanisms when all new generations have negative rewards:
(1) \textit{Re-generation:}
We let $\mathcal{M}_{ASP}$ generate $2 \times N$ additional alternatives if all initial $N$ alternatives for the current input were judged negative by $f_r$.
(2) \textit{Backtracking:} We jump back to the previous hint with maximum reward that had more than one alternative and continue with this as our new partial encoding.
We then restart to generate all successive hints.

\section{Experimental Setup}
\label{sec:six}

\subsection{Evaluation Metrics and Datasets}

We report the accuracy of the models based on how often their output exactly matches the correct answer.
ASP encodings that produce more than one solution in the end are considered wrong in our strict evaluation setup as the considered problems only have unique solutions.
One issue when converting the problem into ASP is the ambiguity of string representation in the answer sets and the evaluation with accuracy when comparing to the ground truth solution (e.g., spaces, underscores, and other ambiguities).
To automatically evaluate the predicted answer sets, we implement a \textit{Levenshtein heuristic} that acts as fallback for fuzzy matching if the answer set does not exactly match the ground truth representation.

We work with \textbf{LogicPuzzles} \citep{mitra-baral-2015-learning} and \textbf{GridPuzzles} \citep{tyagi-etal-2024-step}.
LogicPuzzles is a collection of 124 grid-based puzzles (50 train and 74 test).\footnote{We noticed that in the official dataset release, 26 puzzles from the train set are also part of the test set. In order to evaluate the trained models on unseen data only, we remove these 26 instances from our test set. For comparability with prior work, we also report the evaluation on the full data in the appendix.} %
All puzzles are of size $3 {\times} 4$, i.e., there are 3 entity types, each with 4 instances (e.g., 4 dogs, 4 owners, 4 countries) where each entity must be assigned once.
This results in 4 triples representing a unique solution 
(e.g., each dog is assigned to a different owner from a different country).
GridPuzzles introduces different puzzle sizes ($3{\times}4,~3{\times}5,~4{\times}4,~4{\times}5,~4{\times}6$) and difficulty levels (easy, medium, hard).
We use this dataset only for evaluation.

\subsection{Models}
We train and evaluate four open-weight models from two model families: Llama-3.3 70B and Llama-3.1 8B\footnote{At the time of publication, there is no Llama-3.3 8B model.} \citep{grattafiori2024llama3herdmodels} as well as Qwen3 32B and Qwen3 8B \citep{qwen3technicalreport}.
We disable the thinking mode introduced in the Qwen3 model family.
For the two bigger models, we sample distinct training data using our proposed method and fine-tune them separately on their own part.
To train the 8B models, we use the SFT data drawn from Llama 3.3 70B as it yields a more diverse range of statements than Qwen3 32B.
Additionally, since these smaller models initially possess almost no ASP coding skills, we supplement the training data by adding the same number of instances from the LLASP dataset \citep{coppolillo2024llasp} on top of our data.
The LLASP dataset is a collection of single-statement building blocks (e.g., transitive closures).

We further compare our ASP models to the distilled variants of DeepSeek-R1 \citep{deepseek}.
They are prompted to perform reasoning without making use of ASP.
We use the prompt setup of \citet{tyagi-etal-2024-step} to perform reasoning without an ASP solver.
While we mainly focus on ASP coding skills, we also investigate how our neuro-symbolic method performs compared to reasoning language models (RLMs) on these logic reasoning tasks.
Finally, we test our new inference method on one of the latest closed-source models, GPT-4.1-mini (version \textit{2025-04-14}) and show that our best-of-N sampling method also improves closed-source models without the need for prior training.

\newcommand{\tabBest}[1]{\underline{\textbf{#1}}}
\newcommand{\tabGood}[1]{\underline{#1}}

\definecolor{lightgray235}{RGB}{235,235,235}
\newcolumntype{g}{>{\columncolor{lightgray235}}c}

\begin{table*}[ht!]
\centering
\footnotesize
\setlength{\tabcolsep}{3.3pt}
\rowcolors{2}{gray!10}{white} 
\begin{tabular}{llccccccccccccc||ccc}
\hiderowcolors
\toprule
\multicolumn{2}{c}{\textbf{Setup}} & \multicolumn{4}{c}{\textbf{Llama-3.3 70B}} & \multicolumn{3}{c}{\textbf{Llama-3.1 8B}} & \multicolumn{3}{c}{\textbf{Qwen3 32B}} & \multicolumn{3}{c||}{\textbf{Qwen3 8B}} & \multicolumn{3}{c}{\textbf{DeepSeek Variants}} \\
\cmidrule(lr){3-6} \cmidrule(lr){7-9} \cmidrule(lr){10-12} \cmidrule(lr){13-15} \cmidrule(lr){16-18}
\multicolumn{2}{c}{} & w/o ASP & Base & SFT & +TT & Base & SFT & +TT & Base & SFT & +TT & Base & SFT & +TT & Llama 70B & Qwen3 32B & Llama 8B \\
\midrule 
\showrowcolors

\multirow{1}{*}{2S} & LogicP. & 27.0 & 31.1 & 55.4 & \tabBest{66.8} & 0.0 & 21.6 & \tabGood{46.8} & 17.6 & 23.0 & \tabGood{50.3} & 0.0 & 24.3 & 43.2 & 59.5 & \tabGood{62.1}& 24.3 \\

& GridP. & 9.1 & 8.0 & 16.8 & \tabGood{23.7} & 0.0 & 8.8 & \tabGood{16.8} & 18.2 & 21.9 & \tabBest{33.9} & 0.0 & 12.4 & \tabGood{21.2} & \tabGood{21.9} & 18.2 & 4.4 \\
\midrule

\multirow{1}{*}{PP} & LogicP. & - & 56.8 & \tabBest{78.4} & - & 0.0 & 0.0 & - & 51.4 & \tabGood{67.6} & - & \tabGood{31.1} & 24.3 & - & - & - & -\\

 & GridP. & - & 33.9 & \tabBest{55.5} & - & 0.0 & 0.0 & - & 27.7 & \tabGood{37.2} & - & 9.1 & \tabGood{16.1} & - & - & - & - \\
\bottomrule
\end{tabular}
    \caption{ %
    Accuracy of all models on both datasets in both the single-prompt parsing (2S) and prompt pipeline (PP) setting. All test-time methods (TT) are averaged across five runs. Underlined scores denote the best score within the category, bold scores the overall best results. Every row corresponds to one combination of prompt setting and dataset.
    }
    \label{tab:grid_based_puzzle_results}
\end{table*}

We train LoRA adapters \citep{hu2022lora} on top of the base LLMs.
We use 2 to 4 Nvidia H200 GPUs using DeepSpeed stage 3 \citep{deepspeed}.
The batch size is set to 4 per GPU used during training and the learning rate is set to $5\mathrm{e}{-5}$ for all models, which is a commonly used value for SFT.
The parameters are kept in bfloat16 format.
LoRA rank $r$ and $\alpha$ are both set to $128$.
All models are trained for at most $10$ epochs to keep computation feasible.

For inference, we use both Nvidia and Intel Gaudi 2 acceletor cards in combination with vllm \citep{kwon2023efficient}.
We use clingo v.5.7.1 \citep{DBLP:journals/corr/GebserKKS17} as ASP solver.
We set $N = 5$ with a temperature of $T = 1.0$ for the test-time experiments to get a higher variety of outputs.
Due to the higher variety of outputs, we average test-time runs over 5 distinct runs in the case of LogicPuzzles.
To ensure computational feasibility, we limit the number of backtracking steps to 5 for LogicPuzzles and to 1 for GridPuzzles.

\subsection{Training Data Statistics} 
We use the train split of LogicPuzzles to generate ASP training data.
We request the LLMs to generate $N = 5$ alternatives per input.
In the case of Llama-3.3 70B, we get $3.7$ chosen responses and $1.3$ rejected responses on average with a standard deviation of $1.82$. 
Qwen3 32B yields $3.3$ chosen and $1.7$ rejected responses on average with a standard deviation of $2.04$.
This indicates that during sampling, Llama produces a slightly higher number of correct instances, whereas Qwen is not yet at the same level of ASP coding skills.
Finally, we end up with $1.546$ training instances from Llama and $1.060$ from Qwen.
Sampling ASP encodings from the 8B models did not yield sufficient results as they initially lack ASP knowledge and hence do not generate useful responses.

\subsection{Settings}
To test the impact of injecting ASP coding knowledge into LLMs, we compare two settings: 
First, we use a \textbf{single-prompt} \textbf{two-shot (2S)} prompt with limited engineering effort to generate the ASP program hint-wise as during training data generation. For this, we provide two dataset-specific examples and explain choice rules and different forms of logical OR constructs in ASP. We randomly select parts of two instances for GridPuzzles as there is no training split available.
Second, we plug our models into an existing \textbf{PromptPipeline (PP)} \citep{ishay2023leveraging} %
consisting of 6 prompts specifically tailored to LogicPuzzles. %
This resembles a major prompt engineering effort for tailored ASP program generation to a particular dataset.
Whereas our two-shot setting directly requests the LLM to generate ASP encodings for each natural language input, the pipeline uses the following six steps: (1) structured generation of constants, (2) formatting constants so that they are suited for ASP syntax, (3) generation of predicates that are used to form the relations between the entities, (4) formulation of the choice rule, (5) rewriting natural language constraints if they contain for example complex logical OR constructs, and (6) translating all natural language constraints into ASP at once.
As only the sixth and last step produces one complete ASP encoding, a fine-grained intermediate analysis using the solver is not possible.

\section{Results and Analysis}
\label{sec:seven}

\subsection{Two-Shot Parsing}
The two upper rows of Table \ref{tab:grid_based_puzzle_results} display our main results on LogicPuzzles and GridPuzzles in the single-prompt parsing setup.
All our SFT-trained models outperform their untrained counterparts on the task of ASP generation, emphasizing the necessity of fine-tuning current LLMs on underrepresented programming languages like ASP.
Especially the 70B Llama heavily benefits from the training with two-digit improvements in all settings.
For both 8B models, which do not possess usable ASP skills initially, we observe a notable increase in performance of approximately 20pp.-25pp. (percentage points) on LogicPuzzles and 10pp. on average on GridPuzzles after SFT-based training on data automatically drawn from Llama-3.3 70B.
This indicates that our pipeline generates data with qualitative training signals such that small LLMs can learn the essence of ASP coding as well.
Similar performance increases can also be observed for the much harder GridPuzzles dataset on which the models were not specifically trained, i.e., our method demonstrates strong transferability to more complex problem settings.

\newcommand{\gcheckmark}{\textcolor{PineGreen}{\checkmark}}

\begin{table}[t]
\centering
\footnotesize
\setlength{\tabcolsep}{3.5pt}
\begin{tabular}{lccccccr}
\toprule
\textbf{Model Variant} & \textbf{SFT} & \textbf{Seq} & \textbf{Reg} & \textbf{Back} & \textbf{N} & \textbf{LogicP.} & $\Delta$ \\
\midrule
\textbf{Llama-3.3 70B}        &              &              &              &              & 1  & 31.0  & --    \\
+SFT                & \gcheckmark   &              &              &              & 1  & 55.4  & 0.0   \\
~ +Seq               & \gcheckmark   & \gcheckmark   &              &              & 5  & 62.7 & +7.3  \\
~ ~ +Reg            & \gcheckmark   & \gcheckmark   & \gcheckmark   &              & 5  & 64.6  & +9.2  \\
~ ~ +Back           & \gcheckmark   & \gcheckmark   &              & \gcheckmark   & 5  & 62.7  & +7.3  \\
~ ~ +Both (TT)     & \gcheckmark   & \gcheckmark   & \gcheckmark   & \gcheckmark   & 5  & 66.8  &  +11.4\\
                    & \gcheckmark   & \gcheckmark   & \gcheckmark   & \gcheckmark   & 10 & 65.1 & +9.7  \\
                    & \gcheckmark   & \gcheckmark   & \gcheckmark   & \gcheckmark   & 25 & \textbf{69.2} & +13.8 \\
\midrule
\textbf{GPT-4.1-mini}        &              &              &              &              & 1  & 39.2  & 0.0    \\
~ +Seq      & & \gcheckmark   &   &   & 5  & 55.4  & +16.2    \\
\bottomrule
\end{tabular}
\caption{Ablation study results on LogicPuzzles. $\Delta$ shows improvement over +SFT (Llama) or the base model (GPT). (\textbf{Reg} = regeneration, \textbf{Back} = backtracking, \textbf{Seq} = best-of-N w/o regeneration and backtracking)}
\label{tab:logicpuzzle-ablation}
\end{table}

\textbf{Effect of Test-time  Search.}
Next, we apply our reward-based search function to all SFT-trained models with $N = 5$.
We observe great improvements for all models over greedy sampling with $T = 0.0$ (and $N = 1$). 
Moreover, our test-time method leads to competitive performance compared to the state-of-the-art RLMs: ASP-tuned Llama models show strong performance when compared to their Deepseek counterparts.
We also observe similar tendencies for GridPuzzles.

Sampling multiple alternatives with a higher temperature and judging them independently using our reward function $f_r$ greatly increases the robustness and reduces the number of semantically wrong (captured by the factor $\frac{1}{M}$) and erroneous partial ASP encodings as seen by the increased accuracy.
This improved performance indicates that there is still insecurity within the model when it comes to ASP coding and that sampling multiple alternatives mitigates this issue.

A detailed study on the different components used in our test-time methods is shown in Table~\ref{tab:logicpuzzle-ablation}.
Re-generation shows slight improvements of 2pp. over basic best-of-5 sampling.
However, the combination of re-generation and backtracking demonstrates a robust improvement by over 3pp.
This underlines the necessity of both fallback mechanisms: Sometimes regenerating $2 \times N$ alternatives is already sufficient to recover from errors, whereas backtracking allows for deeper changes.
Table~\ref{tab:logicpuzzle-ablation} also shows how the number of generations $N$ influences our best-of-N sampling method.
We observe bigger improvements, achieving up to 69\% accuracy for $N = 25$, which shows that there is a valuable trade-off between inference runtime and accuracy on the problems.
Finally, for the closed-source GPT-4.1-mini without any specific fine-tuning for ASP, we still observe a performance improvement of 16.2pp. when used with our reward-based search. This highlights the value of our solver-in-the-loop setup for a wider range of LLMs.

\begin{figure}
    \centering
    \includegraphics[width=1.0\linewidth]{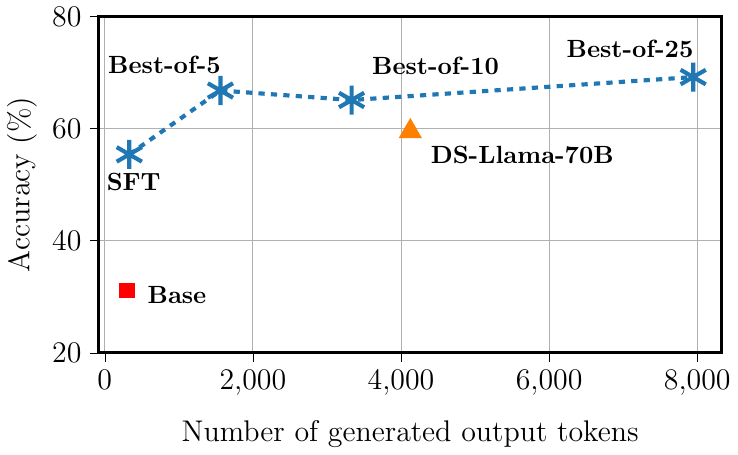}
    \caption{Comparison of generated output tokens in relation to accuracy on LogicPuzzles between the DeepSeek Llama-70B, the untrained Llama-3.3 70B as well as the SFT-tuned one in combination with reward-based inference.}
    \label{fig:output_tokens}
\end{figure}

\textbf{Cost Comparisons.}
Figure \ref{fig:output_tokens} further provides a calculation of the generated output tokens per instance on average in relation to the accuracy on LogicPuzzles for the RLM DeepSeek Llama 70B compared to the ASP-based setups.\footnote{We use the number of generated output tokens as a general proxy for compute costs, as this greatly influences generation speed and costs - two of the main factors when deciding for LLMs.  We do not count input tokens because this is highly problem-specific and depends mainly on the input prompt.}
First, we can see that our trained Llama 70B outperforms the untrained base version by 24.4pp. while keeping the same amount of produced output tokens.
Furthermore, the number of generated output tokens scales linearly with $N$.
We conclude that for increasing values of $N$, the number of produced output tokens is still in a feasible range.
Thorough trade-off considerations for inference costs, however, must be taken into account for determining the best value for $N$.
Finally, we can see that the number of produced output tokens of the RLM is in the range of our best-of-10 model while performing worse on LogicPuzzles.
We can conclude that a value of $N=5$ provides a good trade-off between cost efficiency and task performance in our setup.

\subsection{ASP-tuned Backbones for PromptPipeline}

The bottom row of Table~\ref{tab:grid_based_puzzle_results} shows our ASP models plugged into the 6-step PromptPipeline of \citet{ishay2023leveraging} to test instruction-following capabilities.
We find that fine-tuning the models also improves their instruction following capabilities w.r.t. producing proper ASP encodings.
This shows that ASP-specific fine-tuning using few-shot prompts also generalizes to different settings that follow different approaches by instructing the model to first perform intermediate rewrite steps before finally producing the ASP encoding.

\subsection{Discussion}
One important strength of our method is that it requires no additional human annotations 
as it samples ASP statements from LLMs and automatically computes a reward or categorizes them into \textit{chosen} and \textit{rejected}.
Yet, this also implies that even the chosen instances might not be perfect.
We have shown that recent LLMs, despite possessing limited ASP coding skills, 
are able to produce initial ASP encodings that facilitate our successful self-supervised training data generation method.
Furthermore, our experiments show that best-of-N sampling using a solver-in-the-loop setup mitigates stability issues of the ASP generation process with LLMs by filtering erroneous ASP encodings from a set of $N$ alternatives during inference time.

\section{Related Work}
\label{sec:two}
We review related work on neuro-symbolic systems with a focus on systems in which LLMs generate structured representations and solvers compute solutions \citep{kautz2022third}, ASP code generation, data generation with feedback incorporation, and test-time methods.

\noindent\textbf{Semantic parsing in neuro-symbolic systems.}
Inducing explicit representations of meaning from text relates to the task of \textit{semantic parsing} 
\citep{hendrix1978developing,delmonte1990semantic,baud1998morpho}.
Recently, LLMs have been used for semantic parsing into logic programming languages.
LINC \citep{olausson-etal-2023-linc} translates natural language premises and conclusions into symbolic representations for a theorem prover. 
\citet{schrader-etal-2024-quite} and \citet{nafar2024probabilisticreasoninggenerativelarge} extend this idea to probabilistic reasoning with numeric probabilities and uncertainty. %
\citet{pan2023logic} create symbolic representations comparable to PROLOG and first-order logic and fine-tune LLMs based on error messages from the solver.

\noindent\textbf{ASP code generation with LLMs.}
Early approaches combining NLP methods with ASP propose ideas for solving question answering and reasoning tasks \citep{baral-etal-2004-using,nouioua-nicolas-2006-using}.
The \textsc{Logicia} system \citep{mitra-baral-2015-learning} combines a pairwise Markov network for entity extraction with a maximum entropy model for relation classification on the LogicPuzzles dataset.
\citet{coppolillo2024llasp} introduce the LLASP dataset with single-line building blocks of ASP code.
We focus on deriving ASP encodings for solving entire complex puzzles.
\citet{ishay2023leveraging} create a detailed prompting pipeline for GPT models for solving the LogicPuzzles dataset \citep{brown2020languagemodelsfewshotlearners,openai2024gpt4technicalreport}.
Combining this pipeline with our ASP-tuned models outperforms prior work on GridPuzzles as well.

\noindent\textbf{Generating data from feedback.}
Datasets can %
be labeled either manually or in an automated way \citep{xiao2024comprehensivesurveydirectpreference}. 
The HelpSteer dataset \citep{wang2023helpsteer,dong2023steerlm} is an example for human annotations, while \citet{2023vlfeedback} employ GPT-4V to automatically judge the outputs of vision-language models.
\citet{lai2024stepdpostepwisepreferenceoptimization} use GPT-4 to automatically identify faulty steps in mathematical reasoning chains.
We are not aware of any prior work generating ASP training data from a solver-in-the-loop setup.

\noindent\textbf{Test-time methods.}
Recent test-time methods fall into three categories \citep{dong2024survey}:
\textit{Independent self-improvement} refers to intervening in the generation process of frozen-parameter LLMs \citep{lu-etal-2022-neurologic,ning2024skeleton}.
\textit{Context-aware self-improvement} describes adaptations to prompts by enriching the input to the model, e.g., via chain-of-thought prompting \citep{NEURIPS2022_9d560961}.
Methods using feedback from additional expert \citep{zeng-etal-2024-improving} or reward models \citep{deng-raffel-2023-reward} are called \textit{model-aided self-improvement methods}.
Our approach belongs to the third category.

\section{Conclusion and Outlook}
\label{sec:eight}

In this paper, we have presented a novel method for increasing the performance of LLMs on the task of ASP code generation.
We have shown that an external ASP solver can be used to generate high-quality training data for instruction tuning in a fully automated fashion as well as to provide guidance during inference to filter out faulty ASP encodings.
We observe consistent improvements for multiple models after training in two distinct prompting settings, demonstrating that our training method integrating solver feedback scales to a wider range of LLMs on assignment problems of different sizes and difficulties.
Furthermore, we showed that solver feedback incorporated during test-time can further improve the robustness of the ASP generation process as it is used to find the best combination of partial encodings by ranking them against each other and allows to recover from generation errors.

\paragraph{Outlook.}
Potential next steps involve investigating further training setups.
The usage of reinforcement learning-based (RL) training of LLMs has become popular due to its promising training results \citep{ouyang2022traininglanguagemodelsfollow}, including RL with our solver feedback as a training signal \citep{jha2024rlsf}.
We propose to extend our reward function $f_r$ by introducing weights that rank each error type by its importance.
It can be further extended to also include signals during training that indicate the correctness based on the ground truth solution (similar to our sampling procedure).

\section*{Acknowledgements}
This work was partially supported by the EU Project SMARTY (GA 101140087).

\bibliography{anthology,custom}

\appendix

\section{Explanatory ASP Encoding}
\label{app:explanatory_asp_encoding}
In this section, we provide a full step-by-step explanatory ASP encoding for one test instance of LogicPuzzles.

We take instance $17$ from the test split:

\begin{center}
\fbox{
    \begin{minipage}{0.9\columnwidth}
\textbf{Problem Description:}

MVP Events prides itself on planning large-scale parties for 50 or more people, anywhere in the United States. This month the company has several different events to plan. Using only the clues below, match each event to its number of attendees and the state in which it will be held, and determine which MVP employee is handling the logistics.

\textbf{Entities:}

\textit{people}: 50, 75, 100, 125.

\textit{planners}: Herbert, Joel, Susan, Teresa.

\textit{events}: Anniversary, Birthday, Graduation, Wedding.

\textbf{Clues:}

1. Of the anniversary event and the event with 100 attendees, one will be handled by Joel and the other will be handled by Susan.

2. Herbert's assignment will involve 25 fewer people than Susan's assignment.

3. Of the assignment with 75 attendees and the assignment with 100 attendees, one will be handled by Susan and the other is the birthday.

4. Herbert's event is either the event with 50 attendees or the graduation job.
    \end{minipage}
}
\end{center}

There are three entity types (\textit{people}, \textit{planners}, \textit{events}) with four subjects each.
Hence, it is a $3{\times}4$ grid puzzle.

We begin each ASP encoding by defining all entities and constants that are involved in the search problem.
This is, we start by defining the three predicates \texttt{people}, \texttt{planners/1}, and \texttt{events/1} and assign the 12 entities to each category respectively:

\begin{small}
\begin{Verbatim}[frame=single]
  people(50;75;100;125).
  planners(herbert;joel;susan;teresa).
  events(anniversary;birthday;
         graduation;wedding).
\end{Verbatim}
\end{small}

We could also encode 4 times \texttt{people/1} with a different person inside, but using semi-colons is a much shorter approach.

These 12 entities are now \textbf{hard facts} in the encoding, resulting in them being part of every single possible answer set.

Next, the problem asks for matching each event to its planner and number of attendees.
In ASP, this is called \textit{generate}, i.e., generating potential solutions.
This is achieved by the so-called \textit{choice rule}:

\begin{small}
\begin{Verbatim}[frame=single]
  1 {assignment(Event, Planner, Attendees) 
    : planners(Planner), people(Attendees)} 1 
    :- events(Event).
\end{Verbatim}
\end{small}

The choice rule \textbf{generates} instances of \texttt{assignment/3} with triples (event, planner, attendees).
The syntax \texttt{m {...} n} says that at least $m$ and at most $n$ grounded instances of \texttt{assignment/3} must be generated.
The semantics of this particular choice rule reads as follows: ``For every \texttt{event/1} that is part of the answer set, generate exactly one \texttt{assignment/3} that contains a single event, a single planner, and a single number of attendees.''
Since we defined four distinct events above, the choice rule will generate four grounded \texttt{assignment/3} instances.
However, so far, it does not exclude that planner Herbert, for example, is assigned to two different events, only that every \texttt{assignment/3} will refer to a different \texttt{event/1}.
Therefore, an addition to the choice rule is required, which is formulated as \textbf{rule}:

\begin{small}
\begin{Verbatim}[frame=single]
  {E1 = E2; P1 = P2; A1 = A2} = 0  
  :- assignment(E1, P1, A1), 
     assignment(E2, P2, A2), 
     (E1, P1, A1) != (E2, P2, A2).
\end{Verbatim}
\end{small}

It reads as follows: ``For two \texttt{assignment/3} that are part of the answer set, and the triples (event, planner, attendees) are not exactly the same, there must be zero overlap in any of the three entities event, planner, and number of attendees.''
As we have seen above, there are four \texttt{assignment/3} that are part of the answer set.
This addition now excludes that two distinct events are, for example, assigned the same planner Herbert.

Next, we add all rules and constraints based on the hints provided by the problem instance.

The first clue is \textit{1. Of the anniversary event and the event with 100 attendees, one will be handled by Joel and the other will be handled by Susan.}
It carries multiple implications: The anniversary event does not have 100 attendees and Joel and Susan must plan one of the anniversary event and the event with 100 attendees.
In ASP, this is achieved by the following encoding:

\begin{small}
\begin{Verbatim}[frame=single]
  {E = anniversary; A = 100} = 1  
  :- assignment(E, joel, A).
\end{Verbatim}
\end{small}

This rule says that if there is an \texttt{assignment/3} with Joel, it can only have either the anniversary event or the event with 100 attendees.

Likewise for Susan:

\begin{small}
\begin{Verbatim}[frame=single]
  {E = anniversary; A = 100} = 1  
  :- assignment(E, susan, A).
\end{Verbatim}
\end{small}

The case that both Susan and Joel get the event with 100 attendees assigned is already excluded by the addition to the choice rule described above.
Same holds for the anniversary event.

The next clue (\textit{2. Herbert's assignment will involve 25 fewer people than Susan's assignment.}) is a \textbf{constraint} that excludes a specific condition (as opposed to rules that model if-else statements).
In ASP, this is written as follows:

\begin{small}
\begin{Verbatim}[frame=single]
  :- assignment(_, herbert, A1), 
  assignment(_, susan, A2), 
  not A1 == A2 - 25.
\end{Verbatim}
\end{small}

This is a constraint that requires at least one atom to evaluate to \texttt{false}.
This constraint reads as follows: ``Every answer set that contains one \texttt{assignment/1} for Herbert and one for Susan must not allow for Herbert's number of attendees being anything else than the number of Susan's subtracted by 25.''

The third clue (\textit{3. Of the assignment with 75 attendees and the assignment with 100 attendees, one will be handled by Susan and the other is the birthday.}) is modeled in the same way as clue 1:

\begin{small}
\begin{Verbatim}[frame=single]
  {E = birthday; P = susan} = 1  
  :- assignment(E, P, 75).

  {E = birthday; P = susan} = 1  
  :- assignment(E, P, 100).
\end{Verbatim}
\end{small}

Finally, the fourth clue (\textit{4. Herbert's event is either the event with 50 attendees or the graduation job.}) is an \textit{exclusive-OR} (XOR) that is modeled similarly to clues 1 and 3:

\begin{small}
\begin{Verbatim}[frame=single]
  {E = graduation; A = 50} = 1  
  :- assignment(E, herbert, A).
\end{Verbatim}
\end{small}

The entire encoding looks as follows:

\begin{small}
\begin{Verbatim}[frame=single]
  people(50;75;100;125).
  planners(herbert;joel;susan;teresa).
  events(anniversary;birthday;
         graduation;wedding).

  1 {assignment(Event, Planner, Attendees) 
  : planners(Planner), people(Attendees)} 1 
    :- events(Event).

  {E1 = E2; P1 = P2; A1 = A2} = 0  
  :- assignment(E1, P1, A1), 
     assignment(E2, P2, A2), 
     (E1, P1, A1) != (E2, P2, A2).

  {E = anniversary; A = 100} = 1  
  :- assignment(E, joel, A).

  {E = anniversary; A = 100} = 1  
  :- assignment(E, susan, A).

  :- assignment(_, susan, A2), 
  not A1 == A2 - 25.

  {E = birthday; P = susan} = 1  
  :- assignment(E, P, 75).

  {E = birthday; P = susan} = 1  
  :- assignment(E, P, 100).

  {E = graduation; A = 50} = 1  
  :- assignment(E, herbert, A).

\end{Verbatim}
\end{small}

Running clingo on this encoding returns the following uniquely determined answer set:

\begin{small}
\begin{Verbatim}[frame=none]
    planners(herbert) planners(joel) 
    planners(susan) planners(teresa)
    people(50) people(75) 
    people(100) people(125) 
    events(anniversary) events(birthday) 
    events(graduation) events(wedding) 
    assignment(anniversary,susan,75) 
    assignment(wedding,herbert,50) 
    assignment(birthday,joel,100) 
    assignment(graduation,teresa,125)
\end{Verbatim}
\end{small}

By comparing all four instances of \texttt{assignment/3} to the clues, we can see that this answer set is indeed a stable model of the problem that fulfills all constraints.

\section{Levenshtein Heuristics}
\label{app:levenshtein}
Since exact string matching to compare ground truth and ASP output is often not possible, we implement a Levenshtein heuristic that automatically detects whether an ASP output corresponds to the ground truth or not.
To achieve that, we use the Levenshtein string edit distance that measures how many atomic string edit operations on a character-level (insert, delete, replace) are necessary to transform one string into another.

We want to explain this using an example first.
Consider the following solution of instance with ID $2$ from the test\_HA split, i.e., the split without explanations of how to arrive at the correct solution, of LogicPuzzles:

\begin{small}
    \begin{verbatim}
    (2016, ISON-X42, Dr. Golden)
    (2017, Egert Facility, Dr. Owens)
    (2018, Zynga Complex, Dr. Weber)
    (2019, Bale-Hahn SSC, Dr. Farley)
    \end{verbatim}
\end{small}

Running clingo on the ASP encoding parsed by Llama-3.1 70B yields the following answer sets:

\begin{small}
    \begin{verbatim}
    assignment(ison_x42,golden,2016) 
    assignment(bale_hahn_ssc,farley,2019)
    assignment(egert_facility,owens,2017) 
    assignment(zynga_complex,weber,2018) 
    \end{verbatim}
\end{small}

We can see that the main differences are dashes and spaces being converted into underscores, as well as some shortenings such as the prefix \enquote{Dr.} being removed.
These differences make it impossible to perform direct string comparisons.
Therefore, for comparing computed output and ground truth, we first transform the ground truth into a representation as it could be used in ASP encodings.
However, if comparison still fails, we apply our Levenshtein heuristics to compare both sets.
This heuristic applies the following steps:
\begin{enumerate}
    \item For each computed set of assignments, iterate over each ground truth tuple and every item contained in it and compare it to every single item in the computed answer sets. In this example, we compare every item out of the 12 ground truth entities to all 12 computed items. This results in a runtime complexity of $\mathcal{O}(n^2)$, with $n = 12$ in this running example.
    For example, taking \texttt{ison\_x42} and comparing it to \texttt{(2016, ISON-X42, Dr. Golden)} results in the following three edit distances computed by NLTK's implementation of Levenshtein:
    
\begin{small}
\begin{verbatim}
    edit_distance("ison_x42", 
                  "2016") = 8
    edit_distance("ison_x42", 
                  "ISON-X42") = 6
    edit_distance("ison_x42", 
                  "Dr. Golden") = 10
\end{verbatim}
\end{small}

    After each comparison, we store the index tuple $(i, j), i, j \in [1,\dots,4]$ where $i$ and $j$ refer to the two row indices where the edit distance was the lowest for $i$.
    In the case above, we would have $(1,1)$ unless there is another row where \texttt{ison\_x42} needs less edits.
    However, if there are minimum edit distances, we perform an additional step that checks whether two strings are contained in each other or not.
    If so, this match will be preferred over the match without overlapping strings.

    \item To judge whether a computed solution matches its ground truth counterpart, we check if for every item in ground truth row $i$ the matching row $j$ is the same. Furthermore we require $j$ to be assigned to only one row $i$. For example, the following Levenshtein matrix indicates a valid solution for the running example:

        \begin{small}
        \begin{verbatim}
    [[(1,1), (1,1), (1,1)],
     [(2,3), (2,3), (2,3)],
     [(3,4), (3,4), (3,4)],
     [(4,2), (4,2), (4,2)]]
        \end{verbatim}
    \end{small}

\end{enumerate}

\section{Number of Possible Solutions for Grid-based Puzzles}
In this section, we want to provide a visual explanation for why the number of possible solutions of an $m \times n$ assignment equals to $(n!)^{(m-1)}$, i.e., for a puzzle with $m$ entity categories with $n$ entities in each category.

Exemplarily, we use a $3 \times 4$ grid puzzle.
That could be matching four dogs with four owners and four houses.

\begin{figure}
    \centering
    \includegraphics[width=0.9\linewidth]{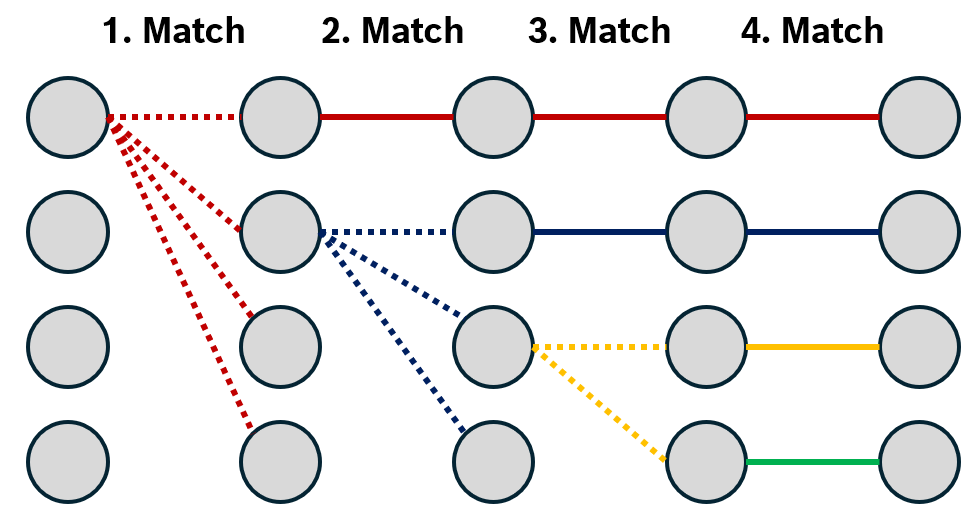}
    \caption{Matching two types of entities with four subjects each. There are $4 \times 3 \times 2 \times 1 = 4!$ possibilities to find exclusive matchings.}
    \label{fig:graph_matching_1}
\end{figure}

We approach this combinatorial problem as graph as shown in figure~\ref{fig:graph_matching_1}.
We start by matching the first $4$ entities with another $4$ entities and only allow for $1:1$ matchings.
When selecting the first entity of category 1, there are 4 opportunities to match it with an entity of category two.
Afterwards, when matching the second entity of category 1, there are only 3 entities left from category 3, resulting in only 3 opportunities.
Likewise, the third entity of category 1 has only 2 opportunities left for a matching, while the final one from category 1 has only a single matching possibility.
This results in $4 \times 3 \times 2 \times 1 = 4!$ opportunities.
The general number for $n$ entities per category is therefore $n!$.

\begin{figure}
    \centering
    \includegraphics[width=0.5\linewidth]{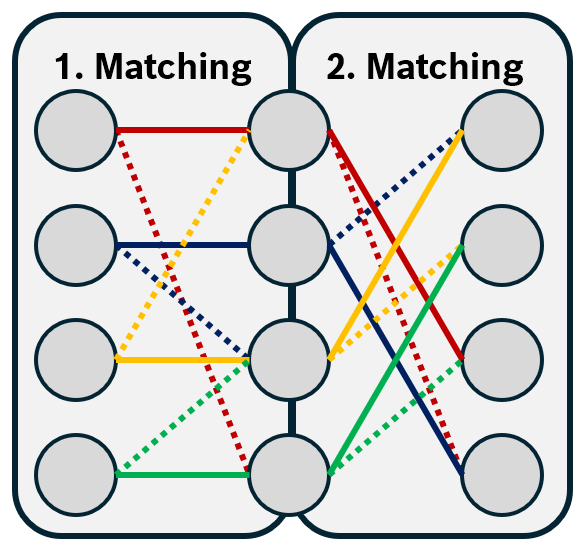}
    \caption{Visualization of an $3 \times 4$ grid puzzles that requires $2 = 3 - 1$ independent matchings with $4!$ possibilities in each subgraph.}
    \label{fig:graph_matching_2}
\end{figure}

The same holds for the second matching between entities from category 2 and category 3.
Hence, we get $2 = 3 -1$ independent matching problems as shown in figure~\ref{fig:graph_matching_2}.
The general formula hence is $m - 1$.
Putting all together results in $(n!)^{(m-1)}$ possible solutions for an unconstrained grid puzzle.

\section{GridPuzzles Example}
GridPuzzles is very similar in its structure to LogicPuzzles.
The main difference is that it goes beyond grid sizes of $3{\times}4$ by additionally providing problems of sizes $3{\times}5$, $4{\times}4$, $4{\times}5$, and $4{\times}6$.

The following instance with ID $8526$ (note that there are still 274 instances) is an example for a $4{\times}4$ puzzle:

\begin{center}
\fbox{
    \begin{minipage}{0.9\columnwidth}
\textbf{Problem Description:}

Maurice had several customers in his tattoo parlor today, each of which requested a simple tattoo of their astrological sign. Using only the clues below, match the prices to the options from customers, colors, and zodiac signs. Remember, as with all grid-based logic puzzles, no option in any category will ever be used more than once.

\textbf{Entities:}

\textit{prices}: \$35, \$40, \$45, \$50.

\textit{customers}: Bonita, Carole, Kendra, Neil.

\textit{colors}: black, pink, red, violet.

\textit{zodiac signs}: Pisces, Sagittarius, Taurus, Virgo.

\textbf{Clues:}
1. Bonita was the Taurus.

2. Of the person who paid \$50 and the Virgo, one got the pink tattoo and the other got the violet tattoo.

3. The Taurus was either the customer who got the red tattoo or the customer who got the violet tattoo.

4. Kendra was either the person who paid \$50 or the Pisces.

5. Of the customer who paid \$35 and Neil, one got the red tattoo and the other was the Pisces.

6. Neil paid 10 dollars more than the customer who got the black tattoo.
    \end{minipage}
}
\end{center}

We approach this dataset in the same way as LogicPuzzles, i.e., we start by defining all the constants:

\begin{small}
\begin{Verbatim}[frame=single]
  price(35;40;45;50).
  customer(bonita;carole;kendra;neil).
  color(black;pink;red;violet).
  zodiac_sign(pisces;sagittarius;
                taurus;virgo).
\end{Verbatim}
\end{small}

Next, we formulate the choice rule.
However, we need to slightly modify it to fit to the $4{\times}4$ instance instead of the $3{\times}4$ ones from LogicPuzzles:

\begin{small}
\begin{Verbatim}[frame=single]
  1 {assignment(P, C, CO, Z) : 
      price(P), customer(C), color(CO)} 1 
      :- zodiac_sign(Z).
      
  { P1 = P2; C1 = C2; CO1 = CO2; Z1 = Z2 } 
  = 0 
      :- assignment(P1, C1, CO1, Z1), 
      assignment(P2, C2, CO2, Z2), 
      (P1, C1, CO1, Z1) != 
            (P2, C2, CO2, Z2).
\end{Verbatim}
\end{small}

The main difference to the choice rule of LogicPuzzles is that we now have 4 entity types in the choice rule.

Next, we look at the first hint: \textit{1. Bonita was the Taurus.}
This is a hard fact. However, in ASP, we cannot add facts with unknown arguments.
However, we only know Bonita being the Taurus, but not how much was paid or what tattoo color there is.
Hence, we need to formulate it as a constraint, thereby filtering out all answer sets generated by the choice rule that do not fulfill this constraint:

\begin{small}
\begin{Verbatim}[frame=single]
  :- assignment(_, bonita, _, Z), 
                Z != taurus.
\end{Verbatim}
\end{small}

The second hint (\textit{2. Of the person who paid \$50 and the Virgo, one got the pink tattoo and the other got the violet tattoo.}) enforces an XOR across two assignments:

\begin{small}
\begin{Verbatim}[frame=single]
  { Co1 = pink; Co2 = pink}  = 1 
    :- assignment(50, _, Co1, _), 
       assignment(_, _, Co2, virgo).
  { Co1 = violet; Co2 = violet } = 1 
    :- assignment(50, _, Co1, _), 
       assignment(_, _, Co2, virgo).
\end{Verbatim}
\end{small}

These two statements enforce pink being assigned to exactly one of both (the \$50 person and the Virgo), as well as violet being assigned to only one of both.
It is not possible that one gets both red and violet in the answer set as this is forbidden by the second statement of the choice rule block above.

The third clue (\textit{3. The Taurus was either the customer who got the red tattoo or the customer who got the violet tattoo.}) is an XOR for one assignment, allowing for only one of two (Taurus being the customer with red tattoo or violet tattoo) to be true:

\begin{small}
\begin{Verbatim}[frame=single]
  { Co = red; Co = violet } = 1 
    :- assignment(_, _, Co, taurus).
\end{Verbatim}
\end{small}

Hint 4 (\textit{4. Kendra was either the person who paid \$50 or the Pisces.}) is modeled the same way as hint 3:

\begin{small}
\begin{Verbatim}[frame=single]
  { P = 50; Z = pisces } = 1 
    :- assignment(P, kendra, _, Z).
\end{Verbatim}
\end{small}

Hint 5 (\textit{5. Of the customer who paid \$35 and Neil, one got the red tattoo and the other was the Pisces.}) is similar to hint 2:

\begin{small}
\begin{Verbatim}[frame=single]
  { Co1 = red; Co2 = red}  = 1 
    :- assignment(35, _, Co1, _), 
       assignment(_, neil, Co2, _).
       
  { Z1 = pisces; Z2 = pisces } = 1 
    :- assignment(35, _, _, Z1), 
      assignment(_, neil, _, Z2).
\end{Verbatim}
\end{small}

Finally, the last hint (\textit{6. Neil paid 10 dollars more than the customer who got the black tattoo.}) is again an ASP constraints that filters all answer sets that do not adhere to this constraint:

\begin{small}
\begin{Verbatim}[frame=single]
  :- assignment(P1, neil, _, _), 
    assignment(P2, _, black, _), 
    P1 != P2 + 10.
\end{Verbatim}
\end{small}

In summary, this is the full ASP encoding for this GridPuzzles instance:

\begin{small}
\begin{Verbatim}[frame=single]
  price(35;40;45;50).
  customer(bonita;carole;kendra;neil).
  color(black;pink;red;violet).
  zodiac_sign(pisces;sagittarius;
    taurus;virgo).

  1 {assignment(P, C, CO, Z) : 
    price(P), customer(C), color(CO)} 1 
    :- zodiac_sign(Z).
  { P1 = P2; C1 = C2; CO1 = CO2; Z1 = Z2 } 
  = 0 
    :- assignment(P1, C1, CO1, Z1), 
      assignment(P2, C2, CO2, Z2), 
      (P1, C1, CO1, Z1) != 
            (P2, C2, CO2, Z2).

  :- assignment(_, bonita, _, Z), 
        Z != taurus.

  { Co1 = pink; Co2 = pink}  = 1 
    :- assignment(50, _, Co1, _), 
    assignment(_, _, Co2, virgo).
  { Co1 = violet; Co2 = violet } = 1 
    :- assignment(50, _, Co1, _), 
    assignment(_, _, Co2, virgo).

  { Co = red; Co = violet } = 1 
    :- assignment(_, _, Co, taurus).

  { P = 50; Z = pisces } = 1 
    :- assignment(P, kendra, _, Z).

  { Co1 = red; Co2 = red}  = 1 
    :- assignment(35, _, Co1, _), 
    assignment(_, neil, Co2, _).
    
  { Z1 = pisces; Z2 = pisces } = 1 
    :- assignment(35, _, _, Z1), 
    assignment(_, neil, _, Z2).
 
  :- assignment(P1, neil, _, _), 
    assignment(P2, _, black, _), 
    P1 != P2 + 10.
\end{Verbatim}
\end{small}

\section{Evaluation on the Full LogicPuzzles Test Split}
\label{app:full_eval}
As already explained above, there is a slight overlap between the train split and the two test splits of the LogicPuzzles dataset.
Therefore, in the main text, we report on the cleaned test split in order to provide a fair evaulation of our models on unseen data.
To also assess the performance of our models on the full test split, we report the numbers calculated on all 100 dataset instances in Table \ref{tab:logicpuzzles_results_all_instances}.
We observe very similar tendencies compared to the cleaned test set, indicating that regardless of the design of the test split, our methods show a clear improvement in terms of ASP coding capabilities.

\begin{table*}[ht!]
\centering
\footnotesize
\setlength{\tabcolsep}{3.3pt}
\rowcolors{2}{gray!10}{white} 
\begin{tabular}{llccccccccccccc}
\hiderowcolors
\toprule
\multicolumn{2}{c}{\textbf{Setup}} & \multicolumn{4}{c}{\textbf{Llama-3.3 70B}} & \multicolumn{3}{c}{\textbf{Llama-3.1 8B}} & \multicolumn{3}{c}{\textbf{Qwen3 32B}} & \multicolumn{3}{c}{\textbf{Qwen3 8B}} \\
\cmidrule(lr){3-6} \cmidrule(lr){7-9} \cmidrule(lr){10-12} \cmidrule(lr){13-15}
\multicolumn{2}{c}{} & w/o ASP & Base & SFT & +TT & Base & SFT & +TT & Base & SFT & +TT & Base & SFT & +TT \\
\midrule 
\showrowcolors

\multirow{1}{*}{2S} & LogicP. & 27.0 & 31.0 & 55.4 & \tabBest{66.8} & 0.0 & 21.6 & \tabGood{46.8} & 17.6 & 23.0 & \tabGood{50.3} & 0.0 & 24.3 & 43.2 \\

\multirow{1}{*}{2S} & LogicP.-\textbf{Full} & 22.0 & 27.0 & 60.0 & \tabBest{68.0} & 0.0 & 30.0 & \tabGood{54.4} & 21.0 & 25.0 & \tabGood{55.6} & 0.0 & 32.0 & 50.6 \\

\multirow{1}{*}{PP} & LogicP. & - & 56.8 & \tabBest{78.4} & - & 0.0 & 0.0 & - & 51.4 & \tabGood{67.6} & - & \tabGood{31.1} & 24.3 & - \\

\multirow{1}{*}{PP} & LogicP.-\textbf{Full} & - & 61.0 & \tabBest{80.0} & - & 0.0 & 0.0 & - & 50.0 & \tabGood{65.0} & - & \tabGood{34.0} & 28.0 & - \\
\bottomrule
\end{tabular}
    \caption{The performance of our trained models compared between the cleaned and the full test split of LogicPuzzles.}
    \label{tab:logicpuzzles_results_all_instances}
\end{table*}

\begin{table}[!ht]
\centering
\begin{tabular}{@{}lcc@{}}
\toprule
\textbf{Training Setting} & \textbf{LogicPuzzles} & \textbf{GridPuzzles} \\
\midrule
SFT & 55.4 & 16.8 \\
DPO & 51.4 & 11.7 \\
\bottomrule
\end{tabular}
\caption{Performance of Llama-3.3 70B on both datasets when trained using either SFT or DPO. We can observe a superior performance of SFT over DPO.}
\label{tab:dpo_vs_sft}
\end{table}

\section{Comparison between DPO and SFT}
Our initial idea was to use pairs of chosen and rejected instances to perform preference alignment-based training using DPO \citep{dpo}.
However, as shown in Table \ref{tab:dpo_vs_sft}, the Llama-3.3 70B model trained with DPO slightly underperforms its SFT-trained counterpart by a few percentage points on both datasets.

\section{Error Type Analysis}
\begin{figure}
    \centering
    \includegraphics[width=1.0\linewidth]{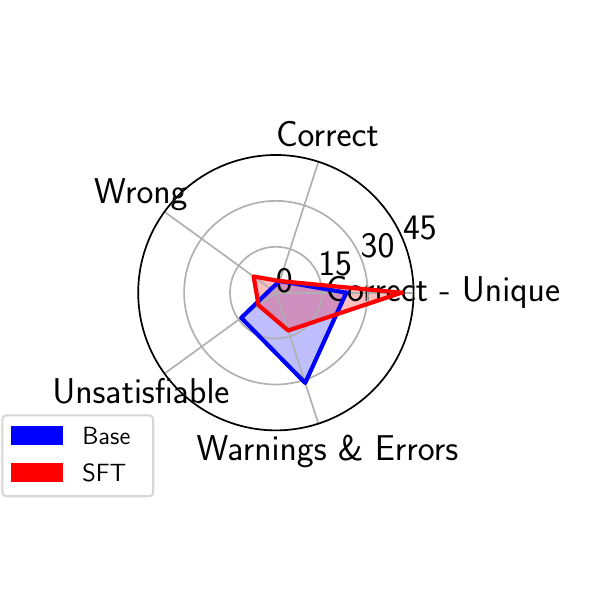}
    \caption{Comparison of success and failure types between the base and SFT trained Llama-3.3 70B.}
    \label{fig:radar_chart}
\end{figure}

Figure \ref{fig:radar_chart} shows the success and error types for the base Llama-3.3 70B compared to its SFT trained counterpart on LogicPuzzles.
We can observe that the untrained Llama especially struggles with producing non-erroneousness and satisfiable encodings.
These cases can be caught especially well in our solver-in-the-loop setup.
Hence, we observe a great shift towards uniquely and correctly answered cases for the SFT trained model.

\begin{figure}[!t]
    \centering
    \includegraphics[width=1.0\linewidth]{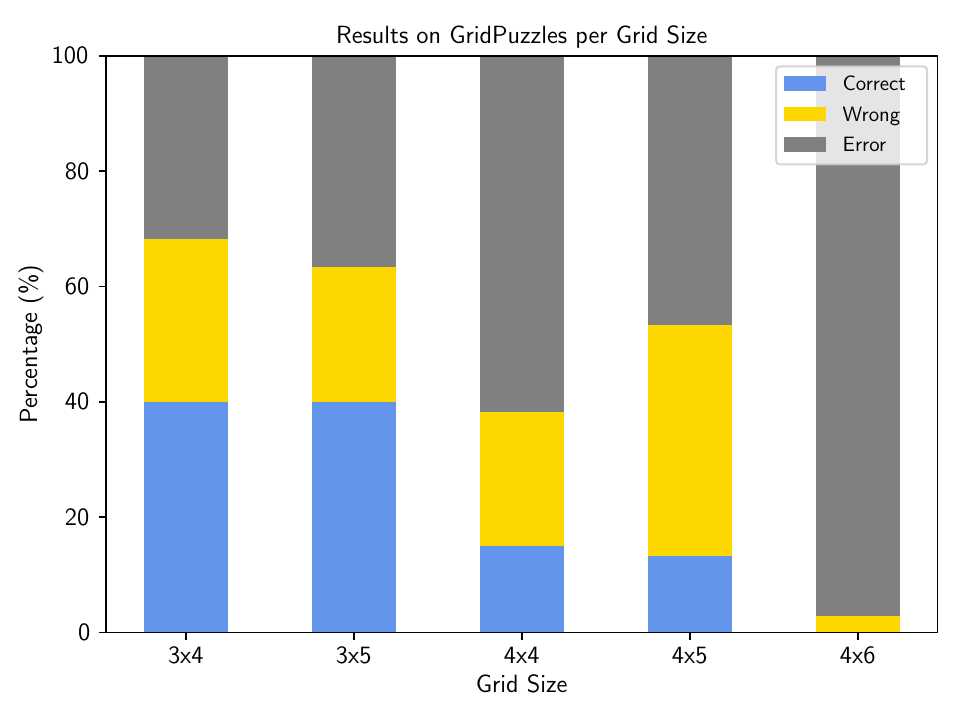}
    \caption{Percentages of correct, wrong, and erroneous cases split by the different grid sizes of GridPuzzles.}
    \label{fig:per_grid_size}
\end{figure}

\begin{figure}[!t]
    \centering
    \includegraphics[width=1.0\linewidth]{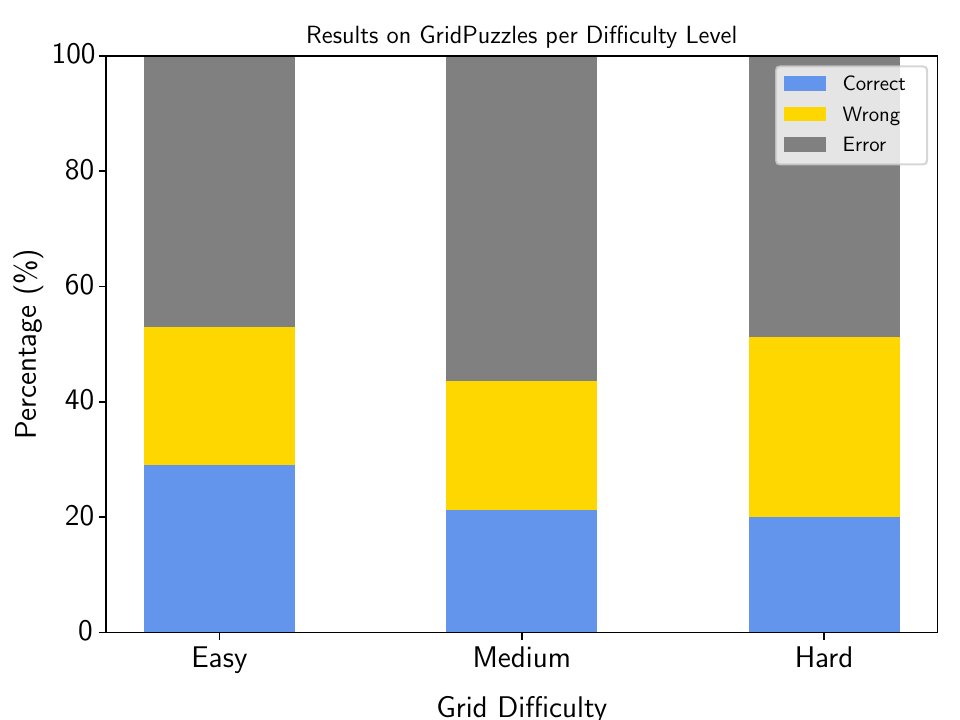}
    \caption{Percentages of correct, wrong, and erroneous cases split by the different difficulty levels of GridPuzzles.}
    \label{fig:per_difficulty}
\end{figure}

\section{Results on GridPuzzles by Size and Difficulty}
GridPuzzles instances are categorized by two dimensions: \textit{grid size} and \textit{difficulty}.
To get a better understanding of how these variables influence the performance of our fine-tuned Llama-3.3 70B in combination with reward-based inference, we plot the percentage of correct, wrong, and erroneous cases for each size and difficulty level in Figures \ref{fig:per_grid_size} and \ref{fig:per_difficulty}.
We can observe that the grid size has a bigger influence on the performance of the system than the human-judged difficulty.
We interpret this as strength of our neuro-symbolic model that it can perform reasoning robustly on even very difficult problems as long as the ASP semantics is correct.
Increasing grid sizes also lead to increasing search spaces for potential solutions, hence requiring much more computational power.
As a result, some of the largest puzzles cannot be solved due to computational constraints.
However, there are strategies for reformulating the uniqueness constraint into $n-1$ separate statements.
Though, in our experiments, the models failed to grasp the semantics of the alternative uniqueness constraints.
We leave this for future research.

\end{document}